# UESA-Net: U-Shaped Embedded Multidirectional Shrinkage Attention Network for Ultrasound Nodule Segmentation

Tangqi Shi[1], Pietro Liò[2]*

## Abstract

**Background:** Breast and thyroid cancers pose an increasing public-health burden. Ultrasound imaging is a cost-effective, real-time modality for lesion detection and segmentation, yet suffers from speckle noise, overlapping structures, and weak global–local feature interactions. Existing networks struggle to reconcile high-level semantics with low-level spatial details. We aim to develop a segmentation framework that bridges the semantic gap between global context and local detail in noisy ultrasound images.

**Methods:** We propose UESA-Net, a U-shaped network with multidirectional shrinkage attention. The encoder–decoder architecture captures long-range dependencies and fine-grained structures of lesions. Within each encoding block, attention modules operate along horizontal, vertical, and depth directions to exploit spatial details, while a shrinkage (threshold) strategy integrates prior knowledge and local features. The decoder mirrors the encoder but applies a pairwise shrinkage mechanism, combining prior low-level physical cues with corresponding encoder features to enhance context modeling.

**Results:** On two public datasets—TN3K (3,493 images) and BUSI (780 images)—UESA-Net achieved state-of-the-art performance with intersection-over-union (IoU) scores of 0.8487 and 0.6495, respectively.

**Conclusions:** UESA-Net effectively aggregates multidirectional spatial information and prior knowledge to improve robustness and accuracy in breast and thyroid ultrasound segmentation, demonstrating superior performance to existing methods on multiple benchmarks.

[1]School of Software Engineering, Xi'an Jiaotong University, Xi'an, 710000, China.

[2]Department of Computer Science and Technology, University of Cambridge, Cambridge, CB3 0FD, United Kingdom.

*Corresponding author. Pietro Liò, Professor, Department of Computer Science and Technology, University of Cambridge, Cambridge, CB3 0FD, United Kingdom; E-mail: pl219@cam.ac.uk.

# 1. Introduction

Over the past decades, the incidence rate of thyroid and breast cancer has been increasing. Among them, breast cancer is one of the three malignant tumors of women, which has become a major public health problem in the current society [1]. Compared with other medical imaging methods, ultrasound can generate images in real time by using the acoustic echo reflected from the interface between different tissues, which has the advantages of low price and minor radiation. It has been recognized as a universal and effective method of medical imaging and preoperative evaluation of breast and thyroid nodules. However, ultrasound examination mainly depends on the judgment of medical experts, and the growth rate of qualified medical experts is far less than that of image data; In addition, the image manifestations of benign and malignant nodules overlap, and the ultrasonic image itself has the disadvantages of high noise and low contrast. In recent years, a series of deep convolution networks have provided excellent performance in ultrasonic image segmentation tasks, such as convolutional neural networks (CNNs), full convolution neural networks (FCNNs) [2] [3], U-Net networks [4] [5] and other neural networks based on encoding and decoding [6]. Ma et al. [7] used the trained CNNs to classify the randomly cut thyroid subgraphs to realize the lesion area segmentation of thyroid ultrasound image. Li et al. [8] used multi-task cascaded FCNNs to segment the focus area of thyroid ultrasound image. The multi-task cascade convolution neural networks (MC-CNNs) proposed by song et al. [9] first locates the lesion area of thyroid ultrasound image, and then uses the deconvolution operation to segment the lesion area in a pyramid manner. Although these methods improve the ability of features to represent the region of interest, the simple feature extraction have difficulty meeting the growing needs of medical applications.

To better improve medical segmentation accuracy, many improved schemes have been introduced, such as squeeze-and-excitation (SE) modules [10]. Considering that the region of interest in medical images may be irregular in shape, which will affect the final segmentation performance, Zhang Y, Yuan L and Wang Y, et al. [11] proposed a spatial attention dense connected U-Net (SAUT-Net) segmentation algorithm, which uses the internal slice attention module to extract the on-chip features, making the spatial information more continuous and accurate. Yamanakkanavar N and Lee B et al. [12] embedded a multiscale strategy and global attention into the SegNet network and proposed a new medical segmentation method, M-SegNet. Xu D, Zhou X and Niu X et al. [13] proposed an automatic segmentation algorithm for low-level glioma images based on U-Net + +. The algorithm normalizes and enhances the samples. Then, feature fusion is performed in the encoding stage. Li C, Tan Y and Chen W et al. [14] combined U-Net++ with an attention network to integrate different levels of feature information. Compared with the original U-Net [15], U-Net++ can effectively improve the segmentation accuracy, but due to the use of dense connections, the number of training parameters of the network increases, resulting in low efficiency. Yan L, Liu D and Xiang Q et al. [16] used a pyramid network for the image segmentation task and learned global features at different scales by grouping the feature maps generated by multiple extended convolution blocks. These do improve the segmentation performance to a certain extent, but they ignore the details of low-level features of ultrasound images and have abundant computing parameters. It is also difficult for these networks to make up for the differences between high-order semantics and low-level features, and they cannot effectively capture the subtle changes in regions of interest. At the same time, there is a lot of overlap in ultrasonic images, which has high noise, and there is weak dependency and interaction between global and local features.

Under the complex conditions of ensuring efficiency and parameter quantity, to strengthen the interaction between local and global semantics, balance the gap between high-level and low-level semantics in the spatial distribution, that is, establish long-term dependence between them, and further improve the segmentation accuracy of ultrasonic images, we propose an ultrasonic image segmentation framework called UESA-Net. The framework is a U-shaped network including two stages of encoding and decoding, which fully captures the fine-grained local details of the lesion area in an ultrasound image and uses low-level information such as a priori physical appearance, size and shape to guide the lesion context and global semantics in detail to obtain multilevel and multiscale complementary information. To some extent, this addresses the deficiency of local semantics in the representation of the lesion area. Besides, considering the spatial complexity of ultrasound images and the differences and mutual characteristics of the internal structure and peripheral tissue of the lesion in the spatial distribution, multidirectional (such as horizontal, vertical and depth) attention networks with a shrinkage strategy are used in both the encoding block and decoding block to model the spatial relationship between the internal structure and peripheral tissue of the lesion area from multiple directions, perform fine processing of features, promote the interaction between them, and highlight the between-class and within-class differences of the internal structure of the lesion area. It is worth noting that the encoding block and decoding block are similar in structure. The difference is that the decoding block adopts the pairwise shrinkage strategy. At the same time, the skip connection operation is used to aggregate the local semantics and prior knowledge of the corresponding encoding layer, which strengthens the spatial dependence between local, global, and low-level prior knowledge and high-order semantics. Finally, the effectiveness and superiority of the proposed UESA-Net segmentation framework are demonstrated on baseline data sets such as TN3K and BUSI. In conclusion, the main contributions of this paper are as follows:

(1) A U-shaped (UESA-Net) ultrasound segmentation framework embedded with multidirectional shrinkage attention is proposed. A novel multidirectional shrinkage attention module is applied to the encoding and decoding path at the same time. The internal structure and peripheral tissue of the lesion area are spatially modelled from different directions and levels to provide fine local details for the lesion area to assist low-level features in obtaining category localization.

(2) Different shrinkage strategies and multilevel feature combinations are used in the encoding and decoding blocks. The fine-grained details captured by multidirectional attention, such as horizontal, vertical and depth information, are shrunk and combined with prior knowledge in the encoding block. In the decoding block, first, the multiscale spatial details captured by multidirectional attention, such as horizontal, vertical and depth information, are combined by the shrinkage strategy. Second, skip connections are used to fuse all the semantic, low-level physical appearance and other prior knowledge obtained by decoding with the local details of the corresponding encoding block to form complementary information. At the same time, low-level physical appearance and other prior knowledge provide detailed guidance for the context and global modelling of the lesion area.

(3) The designed UESA-Net segmentation framework achieves robust segmentation performance on baseline data sets such as TN3K and BUSI and demonstrates the advanced nature of the proposed framework.

The remaining organizational structure of this paper is as follows: The second section describes the data sources, the proposed UESA-Net segmentation framework and its important components in detail. The third section introduces experimental environment, experimental results and ablation analysis. And then in the fourth section, we discuss the issue and describe future research. In the last section, we summarize the full text.

## 2. Materials and Methods

In this section, we introduce our first proposed UESA-Net segmentation framework in detail. Then, we introduce the multidirectional attention encoding module with the shrinkage strategy and the structure and working principle of the decoding module are described in detail.

### 2.A Dataset

To verify the effectiveness of the proposed UESA-Net segmentation framework, a series of experimental evaluations were carried out with baseline data sets such as TN3K and BUSI as experimental samples. The baseline data sets are described as follows:

TN3K [16]: These data include 3493 ultrasound images of thyroid nodules from 2421 patients, in which each ultrasound image contains at least one category of thyroid nodule area, and a large number of color areas are deleted. To ensure the consistency and effectiveness of the experiment, we divided the dataset into three parts: training, validation and testing, including 614 ultrasound images of thyroid nodules and non-thyroid nodules for testing, and the remaining 2879 images were randomly selected from every 5 images as validation data. During training, we fixed the image size to 224*224 and used 5 co-crossovers for validation.

BUSI [17]: Images from female cancer patients were collected with the LOGIQ E9 and LOGIQ E9 agile ultrasound systems in 2018. There are 780 ultrasound images, including 133 normal images, 487 benign images and 210 malignant images. To ensure the consistency of the experiment, these breast ultrasound images were normalized, and the image size was fixed to 256 * 256. In addition, the data set was randomly divided into 80% for the training set and 20% for the test set and remained the same throughout the experiment. At the same time, the segmentation model used 5x cross validation in the training process.

### 2.B Network Architecture

The traditional U-Net [18] [19] has achieved good performance in the task of medical image segmentation by relying on the strong feature encoding and decoding ability, but it is slightly insufficient in the processing of low-level details, the internal structure of lesions and the spatial relationship of peripheral tissues. The UESA-Net segmentation framework proposed in this paper can fully capture their direct spatial relationship. At the same time, prior knowledge, such as low-level physical appearance or size information, can provide effective detailed guidance for the context and global semantics of the lesion area. The segmentation framework is mainly composed of an encoder and decoder of a multidirectional shrinkage attention module. The encoder is mainly used to obtain the local details of the lesion area in the ultrasound image and semantically model it from different directions, different levels and different scales to make full use of the spatial relationship. The decoder uses prior knowledge of low-level physical appearance and size information as guidance in determining basic details and uses the pairwise shrinkage strategy to model the global spatial relationship between the internal structure and peripheral tissue of the lesion. At the same time, skip connections are used to establish a long-term dependency between decoding features and the corresponding encoding features, strengthen the interaction between them, and ensure the differences within and between classes. It is worth noting that the encoder of the UESA-Net segmentation framework is composed of five multiscale and multidirectional shrinkage encoding blocks to make full use of the multiscale spatial information of ultrasonic images. The overall architecture of the UESA-Net segmentation framework is shown in Figure 1.

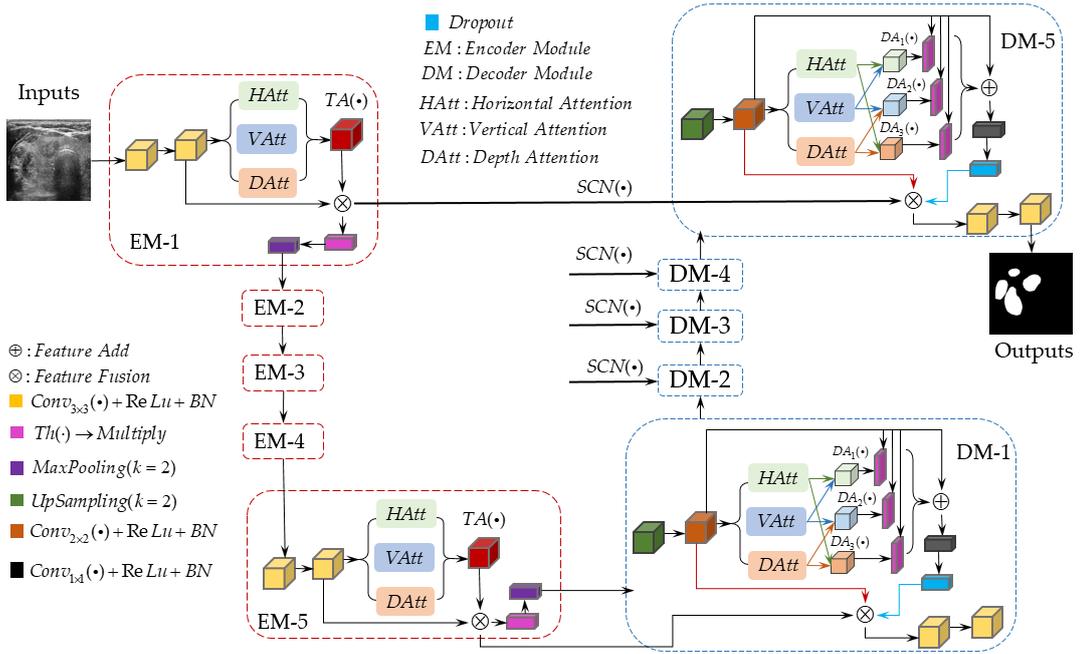

Figure 1: Overall architecture. $TA(\cdot)$ represents an attention fusion module in three directions: horizontal, vertical and depth. $DA_{1\sim3}(\cdot)$ represent three groups of pairwise attention fusion modules. $EM_{1\sim5}, DM_{1\sim5}$ represent five encoding modules and decoding modules with different scales. $SCN(\cdot)$ indicates the skip connection operation.

## 2.B.1 Encoding

According to Figure 1, each encoding block consists of three parts: two consecutive convolution layers [20], a multidirectional attention module with a shrinkage strategy and a maximum pooling layer [21]. After the multidirectional attention module is placed in the continuous convolution layer, the captured feature information is refined from different directions to eliminate redundant information and enhance the feature mapping. At the same time, a spatial relationship is constructed for these local features to effectively resolve the differences in the spatial distribution of the internal structure and peripheral tissues of the lesion.

Suppose the input original image is $x \in R^{W \times H \times C}$, where $W, H, C$ represent the width, height and channel of the input image, respectively. The low-level information captured by the continuous convolution layer is defined as $f_{EM} \in R^{W \times H \times C}$. To capture the spatial relationship more effectively, the low-level feature $f_{EM} \in R^{W \times H \times C}$ is reconstructed, the spatial details are modelled by using multidirectional attention such as horizontal, vertical and depth, and fine processing of local features is performed to further weaken the irrelevant information.

**(1) Horizontal attention module**

First, the low-level feature $f_{EM} \in R^{W \times H \times C}$ is transformed by reshaping to obtain the reconstructed feature $f^1 \in R^{C \times W \times H}$; second, the horizontal attention module is used to establish the local context dependence on the lesion area in a horizontal way, and finally, the horizontal refined feature map $f_{HAtt} \in R^{H \times C \times W}$ is obtained. The operation process of the horizontal attention module is shown in equation (1).

$$f_{HAtt} = HAtt(f_i^1, f_j^1)$$
$$= \sum_{i=1}^{N} \alpha_{ij} x_i$$
$$= \sum_{i=1}^{N} \left( \frac{exp(f_i^1 \cdot f_j^1)}{\sum_{i=1}^{N} exp(f_i^1 \cdot f_j^1)} \right) x_i \qquad (1)$$

where $a_{ij}$ represents the attention coefficient in the horizontal direction. The greater this value is, the stronger the feature correlation in the horizontal direction. $f_i^1, f_j^1$ represents the characteristic diagram of points $i$ and $j$, and $x_i$ represents the elements of the ith characteristic graph.

**(2) Vertical attention module**

First, the low-level feature $f \in R^{W \times H \times C}$ is reconstructed to obtain feature $f^2 \in R^{C \times H \times W}$. Second, the vertical attention module is used to explore the internal structure of the lesion area in a longitudinal way to obtain more refined local semantics and pay greater attention to the key point information. The calculation process of the vertical attention characteristic graph $f_{VAtt} \in R^{C \times H \times W}$ is shown in equation (2).

$$f_{VAtt} = VAtt(f_i^2, f_j^2)$$

$$= \sum_{i=1}^{N} \left( \frac{exp(f_i^2 \cdot f_j^2)}{\sum_{i=1}^{N} exp(f_i^2 \cdot f_j^2)} \right) x_i \quad (2)$$

where N represents the amount of data input to the feature map. The other parameters are the same as in equation (1).

**(3) Deep attention module**

First, the low-level feature $f \in R^{W \times H \times C}$ is transformed and reconstructed, and the obtained feature $f^3 \in R^{W \times C \times H}$ is used as the input information of the deep attention module. Second, the focus spatial semantic information $f_{DAtt} \in R^{C \times H \times W}$ in the depth direction is obtained through the depth attention module, which is similar to the vertical spatial feature $f_{VAtt} \in R^{C \times H \times W}$ in structural form.

To strengthen the interaction between local spatial features in different directions, ensure information complementarity among them, and improve the representation ability of key information in the lesion area, we use the shrinkage strategy to further screen the irrelevant redundant information and better balance the relationship between spatial information in different directions. It is worth noting that in the shrinkage integration stage, a priori knowledge such as low-level physical appearance, size and shape information is introduced to guide the local spatial semantics in detail, which strengthens the correlation and dependence between the internal structure and peripheral tissues of the lesion area.

**(4) Shrinkage integration stage**

According to equations (1) and (2), we obtain spatial details $f_{HAtt}, f_{VAtt}, f_{DAtt}$ in different directions and low-level prior knowledge $f_{EM}$. At the same time, the shrinkage value is defined as Th, and the average feature of multidirectional spatial details such as horizontal, vertical and depth information is $f'_{Avg}$. If the true value in the average feature $f'_{Avg}$ is less than the shrinkage value Th, the "0" tensor generated by the shrinkage is fused with the corresponding prior knowledge $f$ to obtain the fusion feature $f_{TA}$. Otherwise, the "1" tensor generated by shrinkage is fused with the corresponding prior knowledge $f$ to obtain the fusion feature $f_{TA}$. The operation process is shown in equation (3).

$$f_{TA} = \begin{cases} MtP(f_{EM}, T_0), & f'_{Avg} < Th \\ MtP(f_{EM}, T_1), & f'_{Avg} > Th \end{cases} \quad (3)$$

where $MtP(\cdot)$ represents the multiplication of the corresponding elements in the tensor. $T_0, T_1$ represent the "0" and "1" tensors, respectively. The equilibrium characteristic $f'_{Avg}$ is calculated as shown in equation (4).

$$f'_{Avg} = \frac{f_{HAtt} + f_{VAtt} + f_{DAtt}}{3} \tag{4}$$

To model the global features and context dependencies of the lesion area, the fusion feature $f_{TA}$ is maximized. The specific operation is shown in equation (5).

$$f_{mp} = MP_{2\times 2}(f_{TA}) \tag{5}$$

where $f_{mp}$ represents the maximum pooling feature and $MP_{2\times 2}(\cdot)$ indicates the maximum pool operation with a pool core of 2.

The multi-scale image segmentation method proposed by Lian C and Hu x et al. [22] [23] proves that multi-scale information can effectively improve the segmentation performance of lesion region in the image. Therefore, we superimpose multiple encoding blocks to capture multiscale local semantics to enhance the representation ability of encoding features in the lesion region.

## 2.B.2 Decoding

The decoding path is structurally similar to the encoding path. The decoding path mainly includes the upper sampling layer, multidirectional shrinkage attention, convolution layer and skip connections. The feature information extracted from the upper sampling layer is input into the 2 * 2 convolution layer to obtain rich low-level semantics and guide the details of the subsequent global semantics and context information. The multidirectional attentional component adopts a pairwise fusion strategy, which is conducive to obtaining more significant global information. At the same time, to extract rich detail information from ultrasound images and achieve more accurate prediction, skip connections are made between the low-resolution feature map output from the corresponding encoding block, the low-level semantics of the decoding block and the high-resolution global context semantics [24], effective long-term dependence is established between them, the between-class and within-class differences of the internal structure and peripheral tissues of the lesion area are highlighted, and the recognition ability of features with respect to categories is improved. The operation steps of each decoding block are as follows:

**Step 1.** The multiscale spatial information captured by the encoding path is upsampled, and a 2 * 2 convolution layer is input to obtain rich low-level semantics $f_{DM}$.

**Step 2.** The rich low-level semantics $f_{DM}$ are reconstructed to obtain $f_{DM}^1 \in R^{C \times W \times H}$, $f_{DM}^2 \in R^{C \times H \times W}$ and $f_{DM}^3 \in R^{W \times C \times H}$. To eliminate the use of redundant information and strengthen the interaction between the features, the multidirectional shrinkage attention module is used for pairwise fusion, and low-level semantics are introduced in the fusion process to assist the multiscale global information in locating the lesion area. At the same time, the low-level semantics provide detailed guidance for the modelling of global context semantics. In addition, we splice the feature information $f_{DA_1}, f_{DA_2}, f_{DA_3}$ after fusion and shrinkage to form a complementarity relationship to improve the representation ability of global features for the lesion area. It is worth noting that the shrinkage process in the decoding path is similar to that in encoding the road force. The splicing process is shown in equation (6).

$$f_{glob} = Add([f_{DA_1}, f_{DA_2}, f_{DA_3}]) \tag{6}$$

where $f_{glob}$ represents the global semantics of multidirectional shrinkage attention integration in the decoding block. $Add(\cdot)$ represents a simple addition operation.

**Step 3.** The global semantics obtained in step 2 are input into a convolution layer with a size of 1 * 1 for feature mapping and connected with the rich low-level semantics, that is, the prior knowledge and the local semantics output by the corresponding encoding block (skip connections). The operation process is shown in equation (7).

$$\begin{cases} f_{SCN} = SCN(f_{DM}, f_{TA}, f_G) \\ f_G = Conv_1(f_{glob}) \end{cases} \tag{7}$$

where $f_G$ represents the global context semantics after the 1 * 1 convolution, BN layer and dropout operation. $SCN(\cdot)$ indicates the skip connection operation.

This combination of low-level, high-order, local and global contexts makes the segmentation features rich in long-term dependence and low-level physical structure information, and they can accurately segment the focus of ultrasound images. In addition, the superposition of multiple decoding blocks reduces the loss of detail and enhances the generalization of the network.

# 3. Results

This section mainly demonstrates the proposed UESA-Net segmentation framework from the perspectives of data preparation, parameter setting and ablation research and gives the corresponding experimental results and analysis.

## 3.A Experimental Setup

### 3.A.1 Metrics

To ensure the correctness and effectiveness of the experiment, we use the Dice similarity coefficient (DSC), intersection over union (IOU), sensitivity (SEN), specificity (SPE) and accuracy (ACC) to quantitatively analyse the proposed segmentation framework. The specific calculation process is as follows:

$$DSC = \frac{2TP}{2TP+FP+FN} \tag{8}$$

$$IOU = \frac{TP}{TP+FP+FN} \tag{9}$$

$$SEN = \frac{TP}{TP+FN} \tag{10}$$

$$SPEC = \frac{TN}{TN+FP} \tag{11}$$

$$ACC = \frac{TP+TN}{TP+FP+FN+TN} \tag{12}$$

where TP is the true positives; FP indicates false positives, that is, when the true class is predicted to be negative; FN indicates false negatives; and TN indicates true negatives.

### 3.A.2 Parameter settings

To ensure the effectiveness of the experiment, the corresponding initialization parameters are set. For the continuous convolution layer, in the encoder, the size of the convolution core is 3 * 3, and the number of convolution filters is 64-128-256-512. In the decoder, the size of the convolution kernels is 3, 2 and 1, the size of the pooled kernel is 2, and the upsampling scale is 2. For the multidirectional shrinkage attention module, in the encoding stage, we fuse the attention features in the horizontal, vertical and depth directions, while in the decoding stage, we adopt the pairwise fusion strategy; however, the shrinkage value (threshold) Th remains the same, that is, it is controlled within the range of 0.1 to 0.4. It should be noted that when TN3K is used as the training sample, the shrinkage value Th is set to 0.3; when using BUSI as the training sample, the shrinkage value Th is set to 0.2. In the overall training process of the UESA-Net segmentation framework, the initial learning rate is set to 0.001, the dropout rate is set to 0.5, the number of iterations is set to 60 (TN3K) and 100 (BUSI), and the batch size is 6. Finally, to ensure the smooth convergence of the segmentation framework, Adam is used to optimize and adjust the whole framework.

All the experiments in this paper are tested and demonstrated on three V100 GPUs. The main development libraries are Python 3.6, Keras and TensorFlow-GPU, and the other deep learning libraries are NumPy, PIL and Matplotlib.

### 3.B Comparison

To demonstrate the superiority of the proposed UESA-Net segmentation framework, experiments are performed to compare it with many advanced methods, such as AttU-Net [25], ResU-Net [26], SegU-Net [27], TICNet [28] and U-Net [4], and the experimental results and analysis are given. The specific experimental results are shown in Table 1.

Table 1 Experimental results of different segmentation methods. Among them, the TN3K data are iterated 60 times in the training process, while the BUSI data are iterated 100 times. UESA-Net represents the ultrasonic segmentation framework proposed in this paper. "±" indicates the error range. The bold part indicates the optimal segmentation result.

| Datasets | Models | DSC | IOU | SEN | SPEC | ACC |
|---|---|---|---|---|---|---|
| TN3K | AttU-Net[25] | 0.9113±0.0704 | 0.8448±0.1193 | 0.8930±0.0947 | 0.9321±0.0436 | 0.9672±0.1964 |
| | ResU-Net[26] | 0.8576±0.1149 | 0.7685±0.1778 | 0.8237±0.1615 | 0.9026±0.0572 | 0.9501±0.0369 |
| | SegU-Net[27] | 0.8866±0.0819 | 0.8128±0.1354 | 0.8596±0.1048 | 0.9158±0.0569 | 0.9517±0.0234 |
| | TICNet[28] | 0.9112±0.0706 | 0.8446±0.1195 | 0.8913±0.0969 | 0.9339±0.0413 | 0.9672±0.0894 |
| | U-Net[4] | 0.8667±0.0466 | 0.7976±0.0376 | 0.8530±0.0152 | 0.9076±0.0193 | 0.9487±0.0271 |
| | **UESA-Net** | **0.9139**±0.0675 | **0.8487**±0.1151 | **0.9097**±0.0732 | **0.9182**±0.0619 | **0.9678**±0.0051 |
| BUSI | AttU-Net[25] | 0.7092 | 0.6201 | 0.7824 | 0.9649 | 0.9545 |
| | ResU-Net[26] | 0.6487 | 0.5477 | 0.6969 | 0.9651 | 0.9447 |
| | SegU-Net[27] | 0.6597 | 0.5625 | 0.7286 | 0.9607 | 0.9469 |
| | TICNet[28] | 0.7318 | 0.6417 | 0.7687 | 0.9685 | 0.9544 |

| | | | | | |
|---|---|---|---|---|---|
| U-Net[4] | 0.6888 | 0.5965 | 0.7791 | 0.9612 | 0.9522 |
| **UESA-Net** | **0.7344** | **0.6495** | **0.8032** | **0.9693** | **0.9576** |

According to the experimental results in Table 1, we can draw the following conclusions:

(1) The proposed UESA-Net ultrasonic image segmentation framework achieves the best results on the TN3K and BUSI data sets, with IOUs of 0.8487 and 0.6495, respectively. The main reason for these optimal results is that first, we use the U-Net encoder and decoder to model the low-level and high-order information of the lesion area in the ultrasound image, and we use the multiscale strategy in the modelling process to adaptively focus the lesion targets with different scales and shapes. Second, in the multiscale feature combination stage of encoding and decoding, multidirectional shrinkage attention modules such as horizontal, vertical and depth modules are used to refine the features of the lesion area in different directions to locate the key points of the lesion more accurately. At the same time, the use of redundant information is weakened, and the multidirectional attention model also better gathers the context information. It is worth noting that the threshold fusion or screening strategy is equivalent to a gating device to a certain extent, which is more helpful in establishing the long-term dependency between local and global and between low-level and high-level semantics. At the same time, it further strengthens the interaction between different levels of features. In addition, we use attention to fuse encoding features and decoding features, that is, to perform pairwise fusion, which further improves the representation ability of features and reuses the prior information. Therefore, compared with other segmentation methods, our proposed UESA-Net ultrasonic segmentation framework achieves the best performance on multiple baseline data.

(2) Compared with ResU-Net [26] and U-Net [4], SegU-Net [27] achieved better segmentation performance. For example, on the TN3K data, the IOU is 0.0443 and 0.0152 higher than those of ResU-Net [26] and U-Net [4], respectively. It may be that SegU-Net [27] performs connection fusion on the low-level and high-order features captured by the encoder and decoder so that the fused features have a stronger representation ability on the lesion area, but the translation invariance is affected when upsampling the low-level feature map. ResU-Net [26] due to the deepening of the number of network layers, the fitting phenomenon appears, which may make poor perception of detailed information; that is, it easily loses the detailed information of the lesion area during feature capture, resulting in a fuzzy boundary segmentation effect and other problems. In addition, the excellent performance of U-Net [4] in ultrasonic image segmentation further shows the effectiveness of the multiscale strategy, skip connection and other operations.

(3) Compared with other segmentation methods, AttU-Net [25] achieved competitive advantages in these ultrasonic data sets. For example, the IOU and DSC values of AttU-Net [25] are 0.8448±0.1193 and 0.9113±0.0704, respectively. The encoding and decoding structures of AttU-Net [25] and U-Net [4] are basically the same, but the skip connection operation changes. U-Net [4] only uses a simple splicing method to fuse the encoding features and corresponding decoding features, which leads to the repeated use of redundant information. The AttU-Net [25] method uses the attention mechanism to replace the original skip connection operation, fuses the encoding features and corresponding decoding features, further refines the fusion features, and weakens the description of nonkey information in the lesion area, that is, reduces the use of redundant information. This further shows that the attention mechanism is helpful in capturing multiscale information, avoiding acquiring redundant information, and refining features.

To further demonstrate the effectiveness of the proposed UESA-Net segmentation framework, we give the example results of many different segmentation methods. The specific example results are shown in Figure 2.

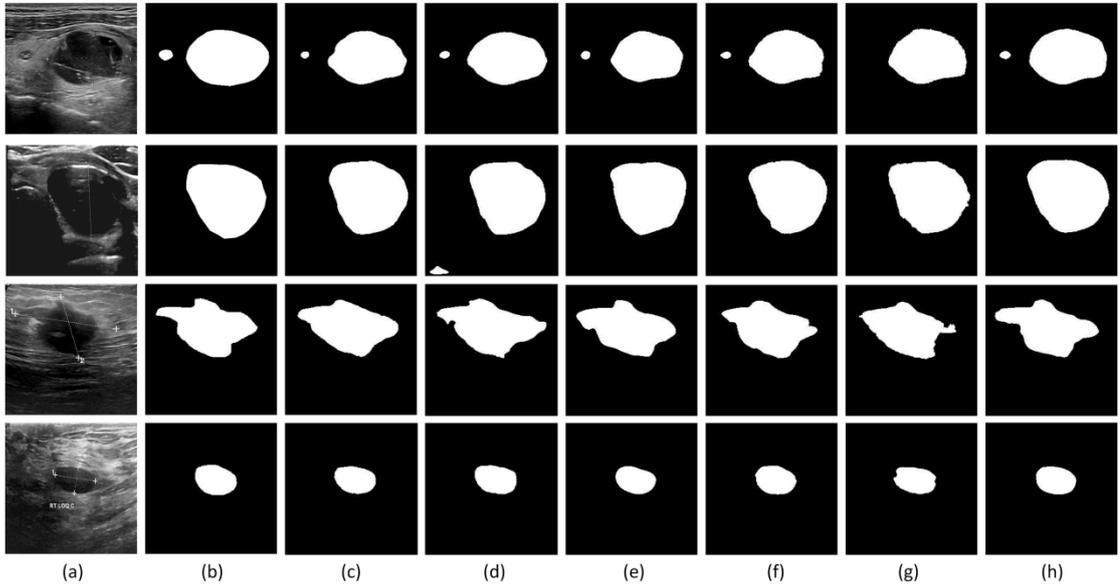

Figure 2: Results of different segmentation methods. (a)Images; (b)Labels; (c)AttU-Net[25]; (d)ResU-Net[26]; (e)SegU-Net[27]; (f)TICNet[28]; (g)U-Net[4]; (h)Ours. The first and second rows are from TN3K; the third and fourth rows are from BUSI.

Compared with the example results of the BUSI baseline data set, the segmentation effect of all methods on the TN3K baseline data set is clearer. In terms of segmentation results, our proposed UESA-Net segmentation framework still has a good visualization effect and is more effective in processing detailed information.

### 3.C Ablation

To demonstrate the impact of different components on the overall performance of the proposed segmentation framework, we use quantitative and qualitative methods to verify the effectiveness of each component on baseline data, such as TN3K and BUSI, to prove its segmentation ability for lesion areas. Next, we will describe the multidirectional attention components and shrinkage strategies in detail.

### 3.C.1 Impact of different thresholds

Taking the above baseline data as experimental samples, the segmentation performance of different thresholds is tested, and the corresponding experimental results and analysis are given. The segmentation performance of different thresholds is shown in Table 2.

Table 2 Segmentation performance of different shrinkage values. Th represents the shrinkage value (threshold) and the value range of Th is 0.1~0.4, that is, $Th = \{0.1, 0.2, 0.3, 0.4\}$. "±" indicates the error range. The bold part indicates the optimal segmentation result.

| Datasets | Models | DSC | IOU | SEN | SPEC | ACC |
|---|---|---|---|---|---|---|
| TN3K | Th=0.1 | 0.9116±0.0725 | 0.8396±0.0957 | 0.8969±0.0634 | 0.9097±0.0721 | 0.9609±0.0068 |
|  | Th=0.2 | 0.9121±0.0589 | 0.8408±0.1233 | 0.8997±0.0529 | 0.9124±0.0841 | 0.9623±0.0042 |
|  | **Th=0.3** | **0.9139±0.0675** | **0.8487±0.1151** | **0.9097±0.0732** | **0.9182±0.0619** | **0.9678±0.0051** |
|  | Th=0.4 | 0.9071±0.0419 | 0.8354±0.1029 | 0.8917±0.0726 | 0.9071±0.0704 | 0.9515±0.0106 |
| BUSI | Th=0.1 | 0.7253 | 0.6318 | 0.7655 | 0.9695 | 0.9553 |
|  | **Th=0.2** | **0.7344** | **0.6494** | **0.8032** | **0.9693** | **0.9576** |
|  | Th=0.3 | 0.7067 | 0.6191 | 0.8077 | 0.9642 | 0.9561 |
|  | Th=0.4 | 0.6823 | 0.5896 | 0.7963 | 0.9627 | 0.9541 |

According to the experimental results in Table 2, we can draw the following conclusions:

(1) For the TN3K baseline data set, the segmentation accuracy is optimal when Th is equal to 0.3, that is, when DSC and IOU are 0.9139±0.0675 and 0.8487±0.1151, respectively, namely, the margin
14

of error remains within 0.12. For the BUSI baseline data, the segmentation effect is the best when Th is equal to 0.2, that is, when DSC and IOU are 0.7344 and 0.6494, respectively. The shrinkage (threshold) strategy in the proposed UESA-Net segmentation framework is helpful in establishing the long-term dependence between local and global as well as low-level and high-level semantic information, and it further highlights the characteristic information of key points in the focus area. For example, the feature information of a key point in the lesion area is similar to that of the peripheral tissue of the lesion. We can highlight the differences between them by using the shrinkage (threshold) strategy, that is, by slicing the similar key area and the peripheral tissue features. If the captured feature information is greater than the shrinkage (threshold), it means that the region is the segmentation key point of the lesion; otherwise, it is the peripheral organization area.

(2) It is obvious from Table 2 that with the increase in the shrinkage (threshold) value Th, the segmentation accuracy first increases and then decreases. For example, taking BUSI as the experimental sample, the segmentation performance of Th = 0.2 is significantly better than that of Th = 0.3; that is, DSC and IOU are improved by 0.0277 and 0.0303, respectively. However, compared with the segmentation accuracy of UESA-Net when Th = 0.3, Th = 0.4 is reduced by 0.0244 (DSC) and 0.0295 (IOU). The main reason is that when the shrinkage (threshold) value is too large, some details are filtered; a larger shrinkage (threshold) will limit the receptive field, and the learned feature information is not sufficient to represent the details of the lesion area. Finally, the representation ability of the lesion information is weakened. When the shrinkage (threshold) is small, irrelevant redundant information may be reused as detail semantics, which increases the sparsity of the network and reduces the segmentation accuracy.

(3) A reasonable setting of shrinkage (threshold) is helpful in improving the image segmentation accuracy, and the setting of the shrinkage (threshold) value depends on the actual situation. For example, Th is set to 0.3 on TN3K baseline data. In summary, a reasonable shrinkage (threshold) setting can effectively improve the segmentation accuracy of the lesion area in the image.

**3.C.2 Impact of attention integration components**

To analyse the contribution of the multidirectional shrinkage attention integration module to UESA-Net, as well as its effectiveness and reliability, an experimental demonstration is carried out on baseline data sets such as TN3K and BUSI, and the experimental results and correlation analysis of different integration strategies are given. It is worth noting that the multidirectional shrinkage attention adopts the pairwise integration strategy in the decoder. The specific experimental results are shown in Table 3.

Table 3 Experimental results of different integration strategies. No-TAtt indicates that the multidirectional shrinkage attention integration module is not embedded in the encoder of UESA-Net. No-DAtt indicates that the pairwise integration strategy of multidirectional attention shrinkage is not adopted in the decoder of UESA-Net. "±" indicates the error range. The bold part indicates the optimal segmentation result.

| Datasets | Models | DSC | IOU | SEN | SPEC | ACC |
| --- | --- | --- | --- | --- | --- | --- |
| TN3K | No-DAtt | 0.9083±0.0734 | 0.8403±0.1237 | 0.8799±0.1112 | **0.9429**±0.0294 | 0.9669±0.0536 |
|  | No-TAtt | 0.9135±0.0685 | 0.8482±0.1165 | 0.8975±0.08974 | 0.9313±0.0455 | 0.9670±0.0034 |
|  | **UESA-Net** | **0.9139**±0.0675 | **0.8487**±0.1151 | **0.9097**±0.0732 | 0.9182±0.0619 | **0.9678**±0.0051 |
| BUSI | No-DAtt | 0.6978 | 0.6084 | 0.7919 | 0.9633 | 0.9553 |
|  | No-TAtt | 0.7177 | 0.6261 | 0.7902 | 0.9667 | 0.9561 |
|  | **UESA-Net** | **0.7344** | **0.6494** | **0.8032** | **0.9693** | **0.9576** |

According to the experimental results in Table 3, we can draw the following conclusions:

(1) Compared with the other two segmentation methods, the proposed UESA-Net segmentation framework still achieves the best results. This is because we use a shrinkage strategy and a multidirectional attention integration module in the encoding and decoding stages to refine the feature information of the lesion area. At the same time, we introduce upper-level prior knowledge to better compensate for the lack of detailed semantics of the local features. In addition, different lesion regions have certain differences in physical appearance and spatial distribution, which are difficult to address with traditional encoding and decoding features. Multidirectional attention and shrinkage strategies can map the feature information of lesion regions into the same subspace and perceive the weak details between them from different directions and angles. They can highlight the between-class and within-class differences to improve the ability of identifying lesion characteristics.

(2) Compared with the No-DAtt segmentation method, the No-TAtt method obtained certain competitive advantages on baseline data sets such as TN3K and BUSI. For example, the IOU evaluation index was increased by 0.0079 (TN3K) and 0.0177 (BUSI), Notably, in the TN3K dataset, the margin of error for IOU remains within 0.15. This shows that embedding the shrinkage strategy and multidirectional attention integration module in the decoder is more conducive to modelling the spatial relationship of the lesion area. The possible reason is that for the attention features in different directions in the decoder, the pairwise integration strategy increases the network's ability to perceive the spatial differences between the lesion area and the peripheral tissue and uses the shrinkage strategy to gather these subtle differences into key features and transmit them to the lower layer. This is complementary to prior knowledge, such as the physical appearance or size of the lower layer and the local semantics of the corresponding encoding layer. At the same time, it establishes an effective long-term dependence to promote the category localization of the lesion area. The No-DAtt segmentation method models only the local spatial information in the encoding stage, ignoring the global and contextual semantics of the lesion area. At the same time, it has poor perception of weak differences within and between classes, which affects the overall performance of the segmentation framework to a certain extent.

(3) The experimental results in Table 3 also show the advantages of multidirectional (such as horizontal, vertical and depth) attention integration components and shrinkage strategies in our proposed UESA-Net segmentation framework. In conclusion, the multidirectional shrinkage attention component helps the proposed UESA-Net segmentation framework perceive the key details of the lesion area, and a series of fine integrations and screenings of coded information and decoded information are carried out in the subsequent feature fusion stage, which eliminates irrelevant redundant information and further improves the feature representation ability of the lesion.

### 3.C.3 Impact of backbone modules

To further demonstrate the effectiveness of the backbone modules of the UESA-Net segmentation framework proposed in this paper, baseline data sets such as TN3K and BUSI are taken as experimental samples, SegNet and U-Net [4] are used as backbone modules for testing and demonstration, and the experimental results and related analysis are given. It is worth noting that these segmentation methods include the multidirectional attention integration module and shrinkage strategy, but the extraction methods of the initial features in the encoder and decoder are different. The experimental results of different backbone modules are shown in Table 4.

Table 4 Experimental results of different backbone modules. Backbone-SegNet means that the backbone module is SegNet. Backbone-UNet indicates that the backbone module is U-Net [4]. "±" indicates the error range. The bold part indicates the optimal segmentation result.

| Datasets | Models | DSC | IOU | SEN | SPEC | ACC |
| --- | --- | --- | --- | --- | --- | --- |
| TN3K | SegU-Net[27] | 0.8866±0.0819 | 0.8128±0.1354 | 0.8596±0.1048 | 0.9158±0.0569 | 0.9517±0.0234 |
| | Backbone-SegNet | **0.8901**±0.0876 | **0.8133**±0.1431 | **0.8677**±0.1178 | 0.9166±0.0534 | **0.9599**±0.0042 |
| | U-Net[4] | 0.8667±0.0466 | 0.7976±0.0376 | 0.8530±0.0152 | 0.9076±0.0193 | 0.9487±0.0271 |
| | Backbone-UNet | 0.8874±0.0902 | 0.8095±0.1468 | 0.8573±0.1309 | **0.9252**±0.0421 | 0.9598±0.0120 |
| BUSI | SegU-Net[27] | 0.6597 | 0.5625 | 0.7286 | 0.9607 | 0.9469 |
| | Backbone-SegNet | 0.6628 | 0.5864 | 0.7724 | 0.9614 | 0.9495 |
| | U-Net[4] | 0.6888 | 0.5965 | 0.7791 | 0.9612 | 0.9522 |
| | **Backbone-UNet** | **0.6949** | **0.6092** | **0.8141** | **0.9623** | **0.9531** |

According to the experimental results in Table 4, we can draw the following conclusions:

(1) Compared with the Backbone-SegNet segmentation method, Backbone-UNet achieves better segmentation performance. For example, on the BUSI baseline data sets, the IOU is increased by 0.0224, respectively. Using U-Net [4] as the backbone module of UESA-Net, the local and global semantics of the lesion area can be modelled by using the encoder and decoder. At the same time, the captured encoding features can be combined with the corresponding decoding layer by using the hop connection operation, which is conducive to the reuse of detailed features and multiscale information. Backbone-SegNet often loses these details.

On the TN3K datasets, Backbone-SegNet segmentation method achieves better segmentation performance on DSC, SEN and IOU, namely, compared to Backbone-UNet, it is improved by 0.0027, 0.0104 and 0.0038 respectively, meanwhile, the error is controlled within the range of 0.1. The mainly reason is that the distribution of positive and negative samples in the TN3K dataset is quite different, and the SegNet backbone module can effectively suppress these unbalanced factors, and at the same time, the fitting phenomenon is alleviated due to the reduction of the number of parameters.

(2) Compared with the SegU-Net [27] and U-Net [4] methods, Backbone-SegNet and Backbone-UNet with the multidirectional attention integration module and the shrinkage strategy achieved advanced performance on TN3K and BUSI samples, and the SPEC values of Backbone-UNet increased by 0.0086 and 0.0009, respectively, compared with those of U-Net [4]. The backbone module of UESA-Net is similar to the traditional U-Net [4] in structure; that is, it includes encoding, decoding, skip connection and other operations. However, we fuse the a priori knowledge many times in the encoding stage and reconstruct the multiscale features of the continuous convolution layer output by using a multidirectional attention and shrinkage strategy. Therefore, the encoder can better grasp the local spatial semantics, multiscale structure information and a priori knowledge and can produce a better model by improving the representation of the key points of the focus area with the encoder, providing more spatial details for the decoder, enhancing the differences between the internal and peripheral tissues of the focus area, and establishing a more effective spatial relationship and long-term dependence.

## 4. Discussion

Researchers could segment high-quality lesion areas of interest by UESA-Net, which provides a solid foundation for ultrasonic study, and fundamentally meets a series of challenges, such as the blur of ultrasonic nodule edge, low background contrast, high noise and so on. Although the proposed UESA-Net segmentation framework achieves good segmentation performance, it is found that the framework still has some shortcomings in the experimental process. For example, the shrinkage value (threshold) needs to be set manually, and the number of model parameters is large. Therefore, in future research, we will develop a lightweight, flexible and adaptable network structure from the perspective of adaptive shrinkage and model simplification, which could further improve the segmentation accuracy.

## 5. Conclusions

In this paper, an ultrasonic image segmentation framework UESA-Net is proposed. Experiments are carried out on ultrasonic images from TN3K and BUSI, and good segmentation results are obtained. For the proposed UESA-Net segmentation framework, first, an encoding and decoding structure similar to that of the traditional U-Net is adopted to model the focus area in the ultrasound image locally, globally, and with low-level and high-order semantics, the improved skip connection is used to effectively address the long-term dependency. Second, the multidirectional shrinkage attention module is integrated into the encoder and decoder to screen and integrate the information flow and to perceive the detailed semantics of the lesion area from different levels and directions. To further compensate for the lack of high-order semantics in the representation of detailed information and the differences in the spatial distribution of the internal structure and peripheral tissues of the lesion, a priori knowledge is introduced into the decoding stage to strengthen the dependency and interaction between local and global and between low-level and high-level semantics. The experimental results show that our proposed UESA-Net segmentation framework has a better segmentation effect on the lesion region in ultrasound images.

**CONFLICT OF INTEREST**

The authors have no relevant conflicts of interest to disclose.